\documentclass{article}

\usepackage{times}
\usepackage{graphicx} 
\usepackage{subfigure} 

\usepackage{natbib}

\usepackage{algorithm}
\usepackage{algorithmic}
\usepackage{hyperref}
\usepackage{booktabs}
\usepackage{natbib}
\usepackage{hyperref}
\usepackage{url}
\usepackage[english]{babel}
\usepackage{natbib}
\usepackage{nicefrac}
\usepackage{algorithm}
\usepackage{algorithmic}
\usepackage{graphics}
\usepackage{graphicx}
\usepackage{tikz}
\usepackage{color,soul,amsmath}
\usepackage{amssymb}
\usetikzlibrary{bayesnet}
\tikzstyle{connect}=[-latex]
\tikzstyle{allconnected}=[line width=0.1cm]
\usepackage{xspace}

\usepackage{url}            
\usepackage{booktabs}       
\usepackage{amsfonts}       
\usepackage{nicefrac}       
\usepackage{microtype}      
\usepackage{algorithm}
\usepackage{algorithmic}
\usepackage{amsmath,amssymb,stmaryrd}
\usepackage{color}
\usepackage{tikz}

\usepackage{xspace}
\newcommand{\acr}[1]{\textsc{#1}\xspace}
\newcommand{\gp}{\acr{gp}}

\newcommand{\bo}{\acr{bo}}

\newcommand{\E}{\mathbb{E}}
\newcommand{\V}{\mathbb{V}}

\newcommand{\II}{\mathbb{I}}

\newcommand{\us}{\acr{pbo}}

\newcommand{\cei}{\acr{cei}}
\newcommand{\ei}{\acr{ei}}

\newcommand{\random}{\acr{ramdom}}

\newcommand{\pe}{\acr{pe}}
\newcommand{\dts}{\acr{dts}}
\newcommand{\ibo}{\acr{ibo}}

\graphicspath{{./diagrams/}}

\global\long\def\inputDim{q}

\global\long\def\dataScalar{y}

\global\long\def\dataSet{\mathcal{D}}

\global\long\def\latentScalar{x}

\global\long\def\latentVector{\mathbf{\latentScalar}}

\global\long\def\kernelScalar{k}

\global\long\def\inputScalar{x}

\global\long\def\inputVector{{\bf \inputScalar}}

\usepackage{color}
\usepackage{verbatim}

\definecolor{brown}{rgb}{0.9,0.59,0.078}
\definecolor{ironsulf}{rgb}{0,0.7,.5}
\definecolor{lightpurple}{rgb}{0.156,0,0.245}

\ifdefined\blackBackground
\definecolor{colorOne}{rgb}{0, 1, 1}
\definecolor{colorTwo}{rgb}{1, 0, 1}
\definecolor{colorThree}{rgb}{1, 1, 0}
\definecolor{colorTwoThree}{rgb}{1, 0, 0}
\definecolor{colorOneThree}{rgb}{0, 1, 0}
\definecolor{colorOneTwo}{rgb}{0, 0, 1}
\else
\definecolor{colorOne}{rgb}{1, 0, 0}
\definecolor{colorTwo}{rgb}{0, 1, 0}
\definecolor{colorThree}{rgb}{0, 0, 1}
\definecolor{colorTwoThree}{rgb}{0, 1, 1}
\definecolor{colorOneThree}{rgb}{1, 0, 1}
\definecolor{colorOneTwo}{rgb}{1, 1, 0}
\fi

\ifdefined\blackBackground

\else

\fi

\global\long\def\cut#1{}

\global\long\def\detail#1{}

\usepackage[accepted]{icml2016}

\begin{document} 

\twocolumn[
\icmltitle{Preferential Bayesian Optimization}

\icmlauthor{Javier Gonz\'alez}{gojav@amazon.com}
\icmladdress{Amazon.com, Cambridge}
\icmlauthor{Zhenwen Dai}{zhenwend@amazon.com}
\icmladdress{Amazon.com, Cambridge}
\icmlauthor{Andreas Damianou}{damianou@amazon.com}
\icmladdress{Amazon.com, Cambridge}
\icmlauthor{Neil D. Lawrence}{lawrennd@amazon.com}
\icmladdress{Amazon.com, Cambridge \\
University of Sheffield}


\vskip 0.3in
]

\begin{abstract}
Bayesian optimization (\bo) has emerged during the last few years as an effective approach to optimizing \emph{black-box} functions where direct queries of the objective are expensive. In this paper we consider the case where direct access to the function is not possible, but information about user preferences is. Such scenarios arise in problems where human preferences are modeled, such as  A/B tests or recommender systems. 
We present a new framework for this scenario that we call \emph{Preferential Bayesian Optimization (\us)} which allows us to find the optimum of a latent function that can only be queried through pairwise comparisons, the so-called \emph{duels}. \us extends the applicability of standard \bo ideas and generalizes previous discrete dueling approaches by modeling the probability of the winner of each duel  by means of a Gaussian process model with a Bernoulli likelihood. The latent preference function is used to define a family of acquisition functions that extend usual policies used in \bo. We illustrate the benefits of \us in a variety of experiments, 
showing that \us needs drastically fewer comparisons for finding the optimum. According to our experiments, the way of modeling correlations in \us is key in obtaining this advantage.

\end{abstract}
\section{Introduction} 
\label{sec:introduction}

Let $g: {\mathcal X} \to \Re$ be a well-behaved \emph{black-box} function defined on a bounded subset ${\mathcal X} \subseteq \Re^{\inputDim}$. We are interested in solving the global optimization problem of finding 
\begin{equation}\label{eq:min_problem}
\latentVector_{min} = \arg \min_{\latentVector \in {\mathcal X}} g(\latentVector).
\end{equation}
We assume that $g$ is not directly accessible and that queries to $g$ can only be  done in pairs of points or \emph{duels} $[\inputVector,\inputVector'] \in  {\mathcal X} \times  {\mathcal X}$ from which we obtain binary feedback $\{0,1\}$ that represents whether or not $\inputVector$ is preferred over $\inputVector'$ (has lower value)\footnote{In this work we use $[\inputVector,\inputVector']$ to represent the vector resulting from concatenating both elements involved in the duel.}. We will consider that $\inputVector$ is the winner of the duel if the output is $\{1\}$ and that $\inputVector'$ wins the duel if the output is $\{0\}$. The goal here is to find $\inputVector_{min}$ by reducing as much as possible the number of queried duels. 

Our setup is different to the one typically used in \bo where direct feedback from $g$ in the domain is available \citep{Jones_2001,Snoek*Larochelle*Adams_2012}. In our context, the objective is a latent object that is only accessible via indirect observations. However, although the scenario described in this work has not received a wider attention, there exist a variety of real wold scenarios in which the objective function needs to be optimized via preferential returns. Most  cases involve modeling \emph{latent human preferences}, such as web design via A/B testing or the use of recommender systems \citep{Brusilovsky:2007}. In prospect theory, the models used are based on comparisons with some reference point, as it has been demonstrated that humans are better at evaluating differences rather than absolute magnitudes \citep{Kahneman79}.

Optimization methods for pairwise preferences have been studied in the armed-bandits context \citep{Yue20121538}. \citet{zoghi:icml14} propose a new method for the K-armed duelling bandit problem based on the Upper Confidence Bound algorithm. \citet{JamiesonKDN15} study the problem by allowing noise comparisons between the duels. \citet{ZoghiKWR15} choose actions using contextual information.
\citet{DudikHSSZ15} study the Copeland's dueling bandits, a case in which a Condorcet winner, or an arm that uniformly wins the duels with all the other arms may not exist. \citet{SzorenyiBPH15} study Online Rank Elicitation problem in the duelling bandits setting. An analysis on Thompson sampling in duelling bandits is done by \citet{DuLS16}. \citet{YueJ11} proposes a method that does not need transitivity and comparison outcomes to have independent and time-stationary distributions.

\begin{figure*}[t!]
\includegraphics[width=1\textwidth]{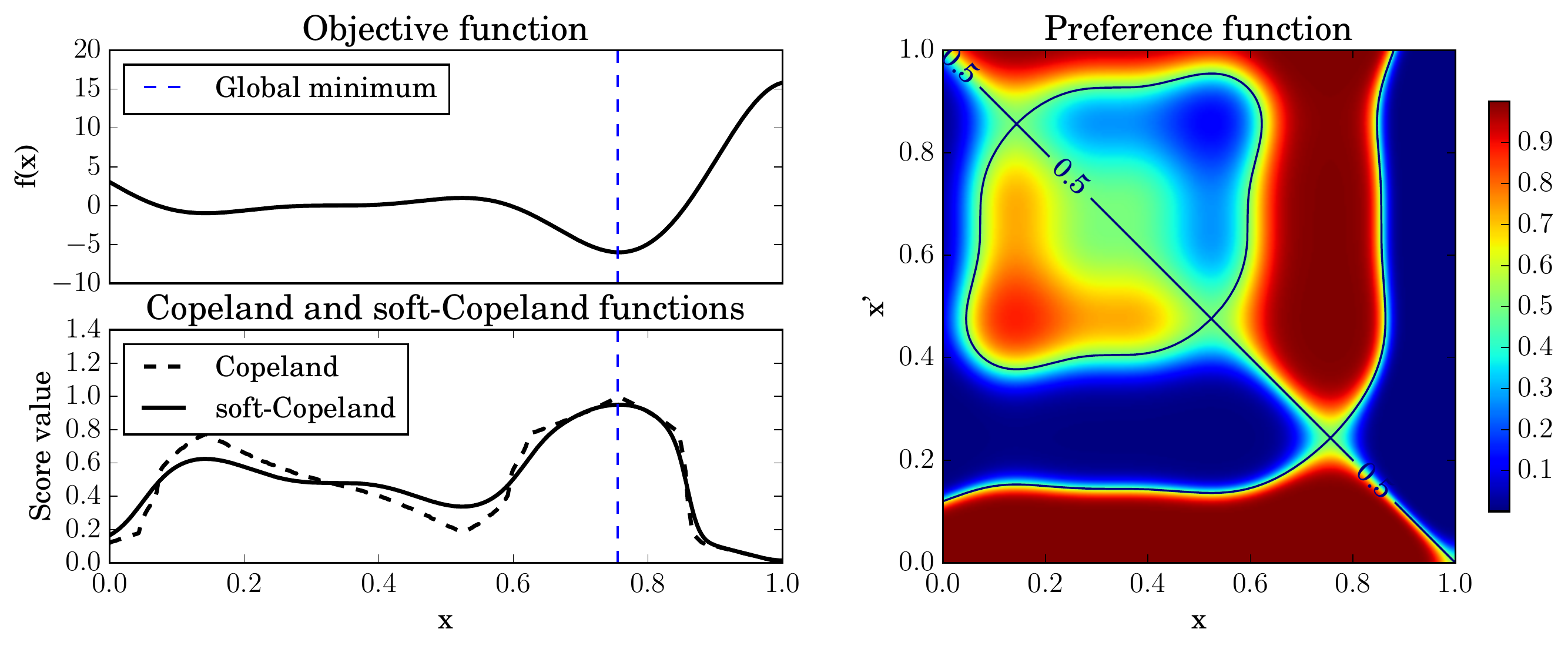}
\vspace{-0.5cm}
\caption{Illustration of the key elements of an optimization problem with pairwise preferential returns in a one-dimensional example. \emph{Top-left}: Objective (Forrester) function to minimize. This function is only accessible through pairwise comparisons of inputs $\latentVector$ and $\latentVector'$. \emph{Right}: true preference function $\pi_{f}([\inputVector,\inputVector'])$ that represents the probability that $\inputVector$ will win a duel over $\inputVector'$. Note that, by symmetry, $\pi_{f }([\inputVector,\inputVector']) = 1-\pi_{f }([\inputVector',\inputVector])$. \emph{Bottom left}: The normalised Copeland's and soft-Copeland function whose maximum is located at the same point of the minimum of $f$.}\label{fig:copeland_score}
\end{figure*}

Preference learning has also been studied \citep{Chu:2005} in the context of Gaussian process (\gp) models by using a likelihood able to model preferential returns. Similarly,  \citet{Brochu:2010} used a probabilistic model to actively learn preferences in the context of discovering optimal parameters for simple graphics and animations engines. To select new duels, standard acquisition functions like the Expected Improvement (\ei) \cite{Mockus77} are extended on top of  a \gp model with likelihood for duels. Although this approach is simple an effective in some cases, new duels are selected greedily, which may lead to over-exploitation.

In this work we propose a new approach aiming at combining the good properties of the arm-bandit methods with the advantages of having a probabilistic model able to capture correlations across the points in the domain ${\mathcal X}$. Following the above mentioned literature in the bandits settings, the key idea is to learn a preference function in the space of the duels by using a Gaussian process. This allows us to select the most relevant comparisons non-greedily and improve the state-of-the-art in this domain.

This paper is organized as follows. In Section \ref{sec:approach} we introduce the point of view that it is followed in this work to model latent preferences. We define  concepts such as the Copeland score function and the Condorcet's winner which form the basis of our approach. Also in Section \ref{sec:approach}, we show how to learn these objects from data. In Section \ref{sec:learning} we generalize most commonly used acquisition functions to the dueling case. In Section \ref{sec:experiments} we illustrate the benefits of the proposed framework compared to state-of-the art methods in the literature. We conclude in Section~\ref{sec:conclusions} with a discussion and some future lines of research.

\section{Learning latent preferences}\label{sec:approach}

The approach followed in this work is inspired by the work of \cite{AilonKJ14} in which cardinal bandits are reduced to ordinal ones. Similarly, here we focus on the idea of reducing the problem of finding the optimum of a latent function defined on ${\mathcal X}$ to determine a sequence of duels on ${\mathcal X} \times {\mathcal X}$.

We assume that each duel $[\inputVector,\inputVector']$ produces in a joint reward $f([\inputVector,\inputVector'])$ that is never directly observed. Instead, after each pair is proposed, the obtained feedback is a binary return $y \in \{0,1\}$ representing which of the two locations is preferred. In this work, we assume that $f([\inputVector,\inputVector']) = g(\latentVector') - g(\latentVector),$ but other alternatives are possible. Note that the more $\inputVector$ is preferred over $\inputVector'$ the bigger is the reward. 

Since the preferences of humans are often unclear and may conflict, we model preferences as a stochastic process.  In particular, the model of preference is a Bernoulli probability function 
$$ p(y = 1 | [\inputVector,\inputVector']) = \pi_{f }([\inputVector,\inputVector'])$$ and 
$$ p(y = 0 | [\inputVector,\inputVector']) = \pi_{f }([\inputVector,\inputVector'])$$
where $\pi: \Re \times \Re \to [0,1]$ is an inverse link function. Via the latent loss, $f$ maps each query $[\inputVector,\inputVector']$ to the probability of having a preference on the left input $\inputVector$ over the right input $\inputVector'$.  The inverse link function has the property that $\pi_{f }([\inputVector',\inputVector]) = 1-\pi_{f }([\inputVector,\inputVector'])$. A natural choice for $\pi_{f}$ is the logistic function
\begin{equation}\label{eq:latent_probability}
\pi_{f}([\inputVector,\inputVector'])  = \sigma(f([\inputVector,\inputVector'])) = \frac1{1+e^{-f([\inputVector,\inputVector'])}},
\end{equation}
but others are possible. Note that for any duel $[\inputVector,\inputVector']$ in which $g(\inputVector) \leq g(\inputVector')$ it holds that $\pi_{f}([\inputVector,\inputVector']) \geq 0.5$. $\pi_{f}$ is therefore a \emph{preference function} that fully specifies the problem. 

We introduce here the concept of \emph{normalised Copeland score}, already used in the literature of raking methods \citep{NIPS2015_6023}, as 
$$S(\inputVector)= \text{Vol} ({\mathcal X})^{-1} \int_{\mathcal X} \II_{ \left\{\pi_{f}([\inputVector,\inputVector'])\geq0.5 \right\}} d\inputVector',$$
where $\text{Vol} ({\mathcal X}) = \int_{\mathcal X} d\inputVector' $ is a normalizing constant that bounds $S(\inputVector)$ in the interval $[0,1]$. If ${\mathcal X}$ is a finite set, the Copeland score is simply the proportion of duels that a certain element $\inputVector$ will win with probability larger than 0.5. Instead of the Copeland score, in this work we use a soft version of it, in which the probability function $\pi_f$ is integrated over ${\mathcal X}$ without further truncation. Formally, we define the soft-Copeland score as
\begin{equation}\label{eq:copeland}
C(\inputVector)= \text{Vol} ({\mathcal X})^{-1} \int_{\mathcal X} \pi_{f}([\inputVector,\inputVector']) d\inputVector',
\end{equation}
which aims to capture the `averaged' probability of $\latentVector$ being the winner of a duel.

Following the armed-bandits literature, we say that $\inputVector_{c}$ is a \emph{Condorcet winner} if it is the point with maximal soft-Copeland score. It is straightforward to see that if $\latentVector_{c}$ is a Condorcet winner with respect to the soft-Copeland score, it is a global minimum of $f$ in ${\mathcal X}$: the integral in (\ref{eq:copeland}) takes maximum value for points $\inputVector \in \mathcal{X}$ such that $f([\inputVector,\inputVector']) = g(\inputVector') - g(\inputVector) >0$ for all $\inputVector'$, which only occurs if $\inputVector_c$ is a minimum of $f$. This implies that if by observing the results of a set of duels we can learn the preference function $\pi_{f}$ the optimization problem of finding the minimum of $f$ can be addressed by finding the Condorcet winner of the Copeland score. See Figure \ref{fig:copeland_score} for an illustration of this property.

\subsection{Learning the preference function $\pi_{f}([\latentVector,\latentVector'])$  with Gaussian processes}\label{sec:model}

Assume that $N$ duels have been performed so far resulting in a dataset $\dataSet = \{[\latentVector_i,\latentVector'_i], \dataScalar_i\}_{i=1}^N$. Given $\dataSet$, inference over the latent function $f$ and its warped version $\pi_f$ can be carried out by using Gaussian processes (\gp) for classification \citep{Rasmussen:2005:GPM:1162254}.  

In a nutshell, a \gp is a probability measure over functions such that any linear restriction is multivariate Gaussian. Any \gp is fully determined by a positive definite covariance operator. In standard regression cases with Gaussian likelihoods, closed forms for the posterior mean and variance are available. In the preference learning, like the one we face here, the basic idea behind Gaussian process modeling is to place a \gp prior over some latent function $f$ that captures the membership of the data to the two classes and to squash it through the logistic function to obtain some prior probability $\pi_f$. In other words, the model for a \gp for classification looks similar to eq.~(\ref{eq:latent_probability}) but with the difference that $f$ is an stochastic process as it is $\pi_f$. The stochastic latent function $f$ is a \emph{nuisance function} as we are not directly interested in its values but instead on particular values of $\pi_f$ at test locations $[\latentVector_{\star},\latentVector_{\star}']$.

Inference is divided in two steps. First we need to compute the distribution of the latent variable corresponding to a test case, $p(f_{\star} | \dataSet, [\latentVector_{\star},\latentVector_{\star}'], \theta )$ and later use this distribution over the latent $f_{\star}$ to produce a prediction
\begin{eqnarray}
\pi_{f}([\inputVector_{\star},\inputVector_{\star}']; \dataSet, \theta)  & = & p(y_{\star} = 1 | \dataSet,  [\inputVector,\inputVector'], \theta) \\ \nonumber
&=& \int \sigma(f_{\star}) p(f_{\star} | \dataSet, [\latentVector_{\star},\latentVector_{\star}'], \theta) df_{\star} \nonumber
\end{eqnarray}
where the vector $\theta$ contains the hyper-parameters of the model that can also be marginalized out. 
In this scenario, \gp predictions are not straightforward (in contrast to the regression case), since the posterior distribution is analytically intractable and approximations at required (see \citep{Rasmussen:2005:GPM:1162254} for details). The important message here is, however, that given data from the locations and result of the duels we can learn the preference function $\pi_f$ by taking into account the correlations across the duels, which makes the approach to be very data efficient compared to bandits scenarios where correlations are ignored.

\subsection{Computing the soft-Copenland score and the Condorcet winner}

\begin{figure*}[t!]
\centering
\includegraphics[width=0.33\textwidth]{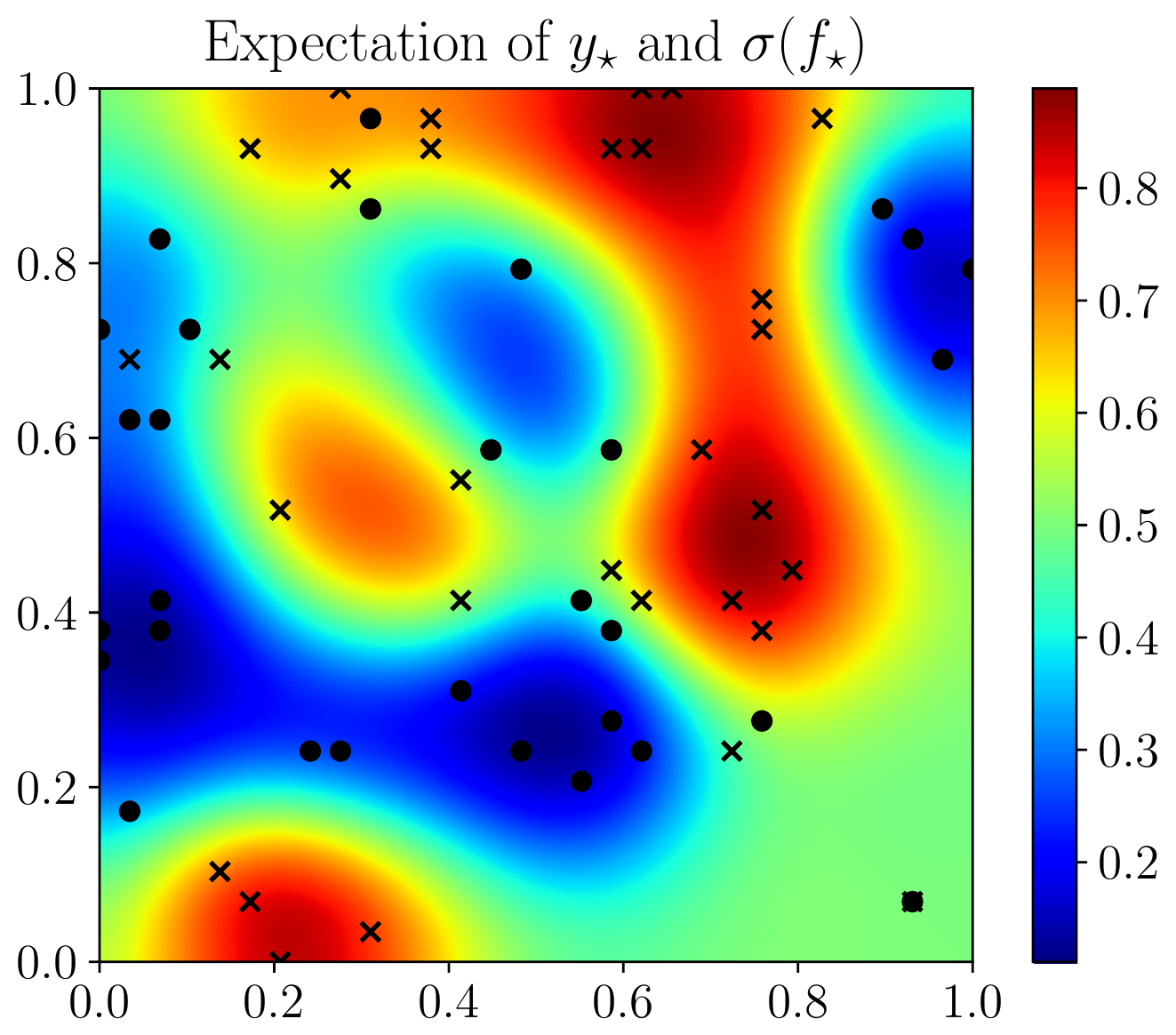} 
\includegraphics[width=0.33\textwidth]{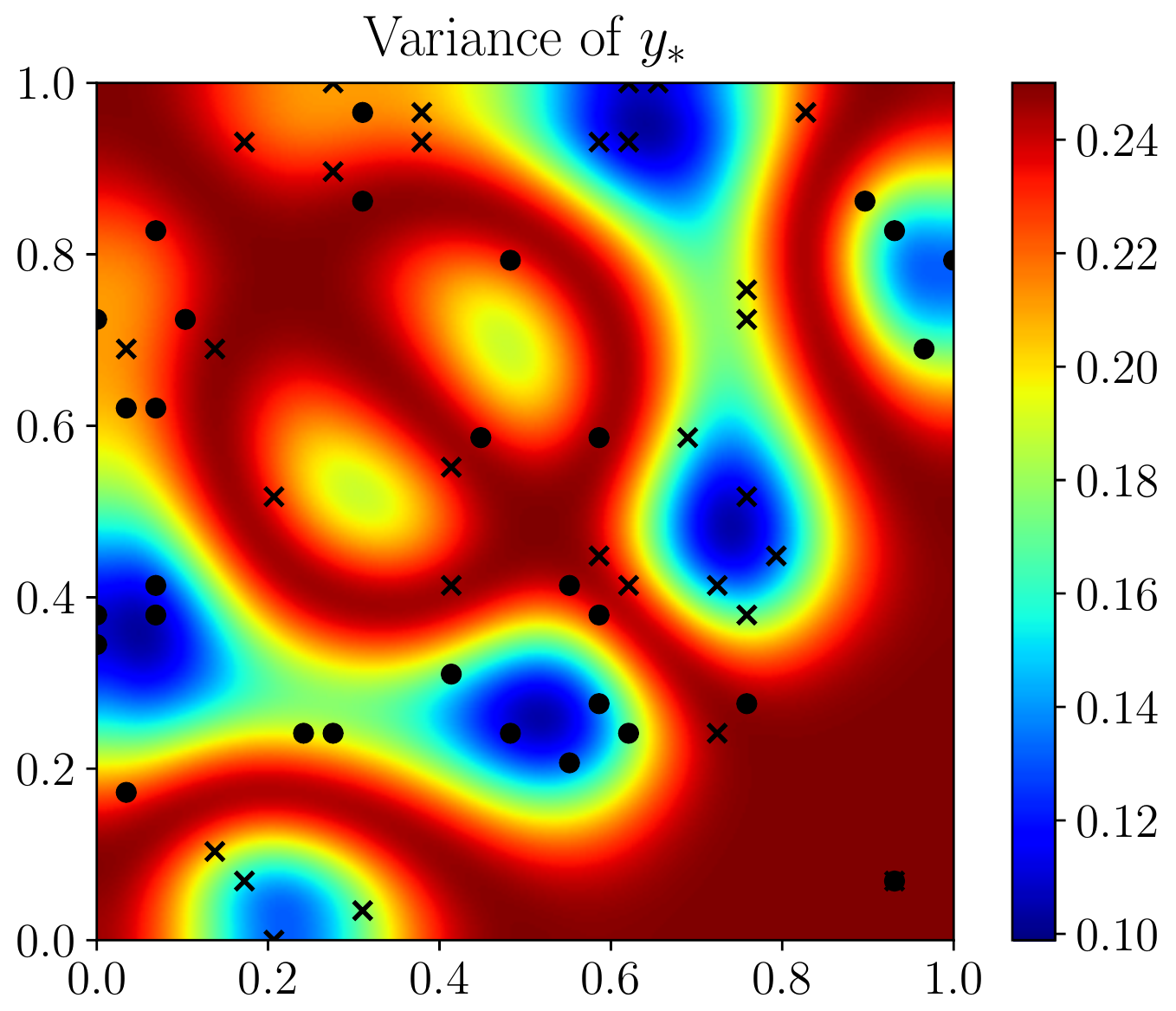} 
\includegraphics[width=0.33\textwidth]{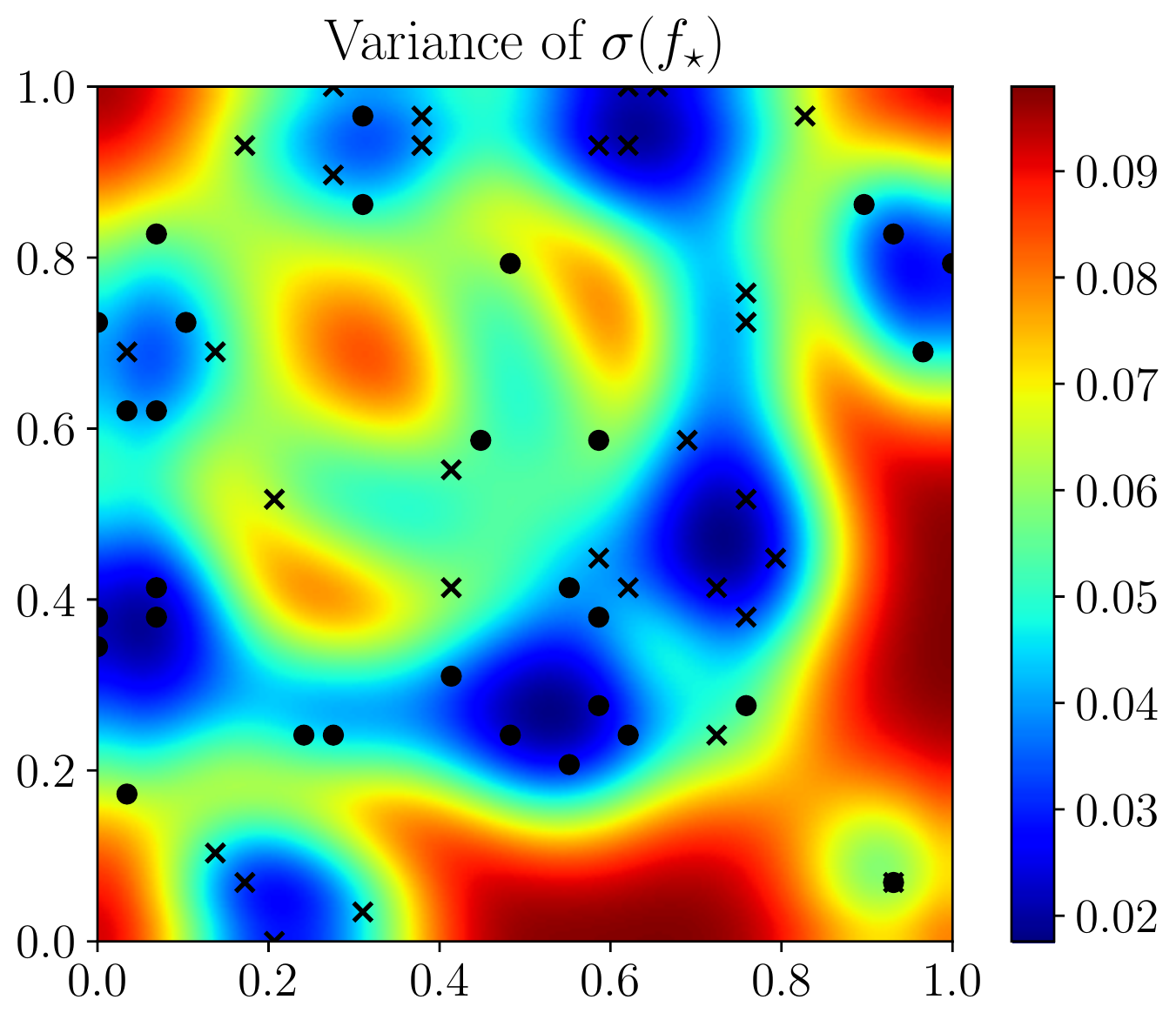}
\vspace{-0.5cm}
 \caption{Differences between the sources of uncertainty that can be used for the exploration of the duels. The three figures show different elements of a \gp 
used for preferential learning in the context of the optimization of the (latent) Forrester function. The model is learned using the result of 30 duels. \emph{Left}: Expectation of $y_{\star}$, which coincides with the expectation of $\sigma(f_{\star})$ and that is denoted as $\pi_{f}([\inputVector_{\star},\inputVector_{\star}']$.  \emph{Center}: Variance of output of the duels $y_{\star}$, that is computed as $\pi_{f}([\inputVector_{\star},\inputVector_{\star}'](1-\pi_{f}([\inputVector_{\star},\inputVector_{\star}'])$. Note that the variance does not necessarily decrease in locations where observations are available. \emph{Right}: Variance of the latent function $\sigma(f_{\star})$. The variance of $\sigma(f_{\star})$ decreases in regions where data are available, which make it appropriate for duels exploration contrast to the variance of $y_{\star}$.} \label{fig:exploration}
\end{figure*}

The soft-Copeland function can be obtained by integrating $\pi_f( [\inputVector,\inputVector'])$ over $\mathcal X$, so it is possible to learn the soft-Copeland function from data by integrating $\pi_f( [\inputVector,\inputVector'], \dataSet)$. Unfortunately, a closed form solution for 
$$\text{Vol} ({\mathcal X})^{-1} \int_{\mathcal X} \pi_{f}([\inputVector,\inputVector']; \dataSet, \theta) d\inputVector'$$ 
does not necessarily exist. In this work we use Monte-Carlo integration to approximate the Copeland score at any $\latentVector \in \mathcal X$ via 

\begin{equation}\label{eq:integral}
C(\inputVector; \dataSet, \theta)  \approx \frac1{M} \sum_{k=1}^M \pi_{f}([\inputVector,\inputVector_k]); \dataSet, \theta), 
\end{equation}
where $\inputVector_1,\dots,\inputVector_M$ are a set of landmark points to perform the integration. For simplicity, in this work we select the landmark points uniformly, although more sophisticated probabilistic approaches can be applied \citep{Briol:2015}. 

The Condorcet winner can be computed by taking
$$x_{c} = \arg \max_{\latentVector \in {\mathcal X}} C(\latentVector; \dataSet, \theta),$$
which can be done using a standard numerical optimizer. $x_{c}$ is the point that has, on average, the maximum probability of wining most of the duels (given the data set $\dataSet$) and therefore it is the most likely point to be the optimum of $g$.

\section{Sequential Learning of the Condorcet winner}\label{sec:learning}

In this section we analyze the case in which $n$ extra duels can be carried out to augment the dataset $\dataSet$ before we have to report a solution to (\ref{eq:min_problem}). This is similar to the set-up in \cite{Brochu:2010} where \emph{interactive} Bayesian optimization is proposed by allowing a human user to sequentially decide the result of a number of duels. 

In the sequel, we will denote by $\dataSet_j$ the data set resulting of augmenting $\dataSet$ with $j$ new pairwise comparisons. Our goal in this section is to define a sequential policy for querying duels:  $\alpha([\latentVector,\latentVector']; \dataSet_j, \theta)$.  This policy will enable us to identify as soon as possible the minimum of the the latent function $g$. Note that here, differently to the situation in standard Bayesian optimization, the search space of the acquisition, $\mathcal{X} \times \mathcal{X}$ is not the same as domain $\mathcal{X}$ of the latent function that we are optimizing. Our best guess about its optimum, however, is the  location of the Condorcet's winner.

We approach the problem by proposing three dueling acquisition functions: (i) pure exploration (\pe), the Copeland Expected improvement (\cei) and duelling-Thompson sampling, which makes explicitly use of the generative capabilities of our model. We analyze the  three approaches in terms of what the balance \emph{exploration-exploitation} means in our context. For simplicity in the notation, in the sequel we drop the dependency of all quantities on the parameters $\theta$ of the model.

\subsection{Pure Exploration}\label{subsec:pureExplore}

The first question that arises when defining a new acquisition for duels, is what \emph{exploration} means in this context. Given a model as described in Section~\ref{sec:model}, the output variables $y_{\star}$ follow a Bernoulli distribution with probability given by the preference function $\pi_f$.
A straightforward interpretation of pure exploration would be to search for the duel of which the outcome is most uncertain (has the highest variance of  $y_{\star}$). The variance of $y_{\star}$ is given by $$\V[y_{\star} | [\latentVector_{\star},\latentVector'_{\star}], \dataSet_j] = \pi_{f}([\inputVector_{\star},\inputVector_{\star}']; \dataSet_j) (1- \pi_{f}([\inputVector_{\star},\inputVector_{\star}']; \dataSet_j)).$$ 
However, as preferences are modeled with a Bernoulli model, the variance of $y_{\star}$ does not necessarily reduce with sufficient observations. For example, according to eq.~(\ref{eq:latent_probability}),  for any two values $\latentVector_{\star}$ and $\latentVector'_{\star}$ such that $g(\latentVector_{\star}) \approx g(\latentVector'_{\star})$, $\pi_f([\inputVector_{\star},\inputVector_{\star}'], \dataSet_j)$ will tend to be close to $0.5$, and therefore it will have maximal variance even if we have already collected several observations in that region of the duels space.

\begin{figure}[t!]
\centering
\includegraphics[width=0.48\textwidth]{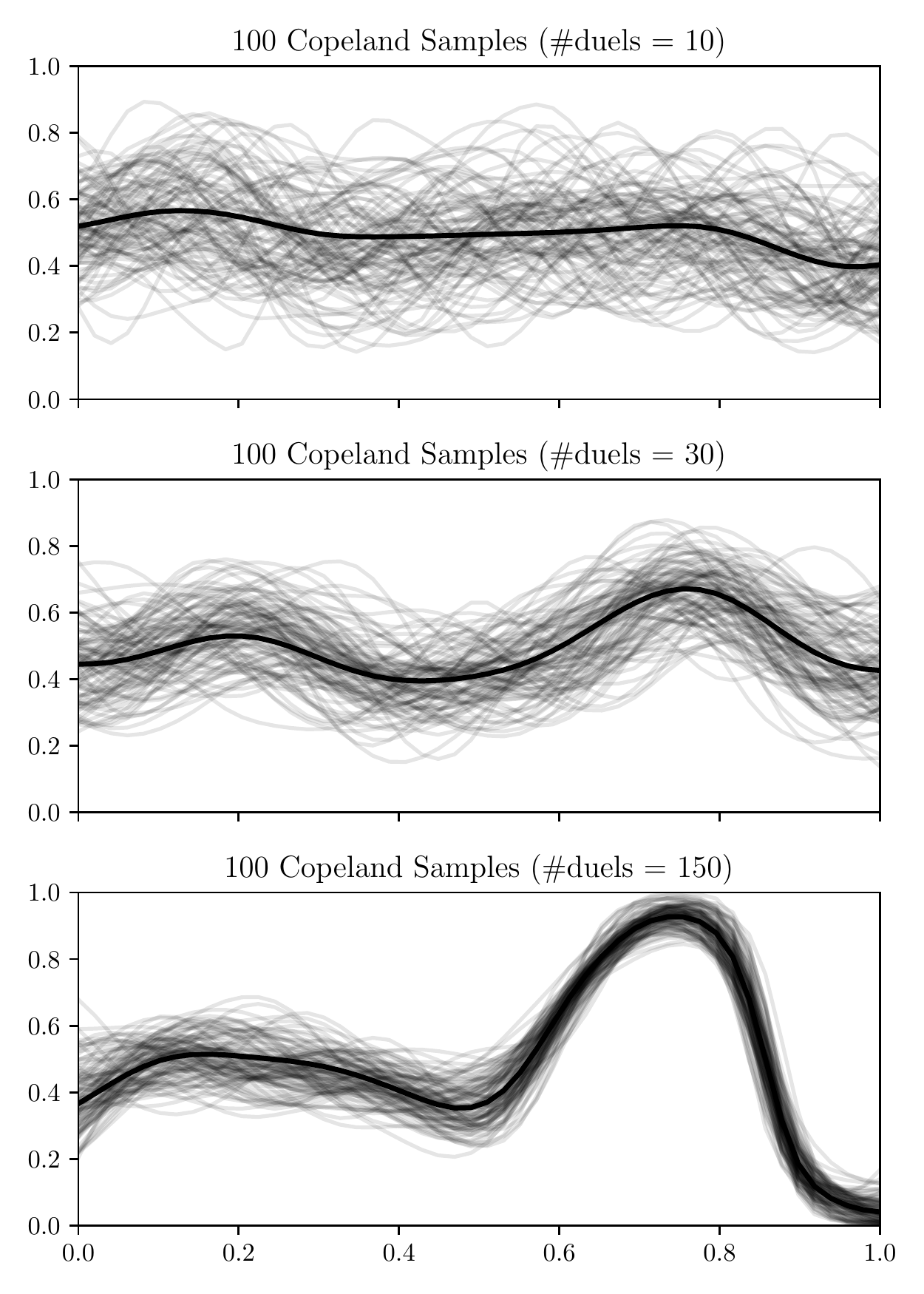}
\vspace{-0.5cm}
\caption{100 continuous samples of the Copeland score function (grey) in the Forrester example generated using Thompson sampling. The three plots show the samples obtained once the model has been trained with different number of duels (10, 30 and 150). In black we show the Coplenad function computed using the preference function. The more samples are available more exploitation is encouraged in the first element of the duel as the probability of selecting $\inputVector_{next}$ as the true optimum increases.}\label{fig:samples}
\end{figure}

Alternatively, exploration can be carried out by searching for the duel where \gp is most uncertain about the probability of the outcome (has the highest variance of $\sigma(f_{\star})$), which is the result of transforming out epistemic uncertainty about $f$, modeled by a \gp, through the logistic function. The first order moment of this distribution coincides with the expectation of $y_{\star}$ but its variance is
 \begin{eqnarray} \nonumber
\V[ \sigma(f_{\star})] & = &\int \left(\sigma(f_{\star}) - \E[ \sigma(f_{\star})] \right)^2 p (f_{\star} | \dataSet,  [\inputVector,\inputVector']) df_{\star} \\  \nonumber
& = &\int \sigma(f_{\star})^2 p (f_{\star} | \dataSet,  [\inputVector,\inputVector']) df_{\star}  - \E[ \sigma(f_{\star})]^2, \nonumber
\end{eqnarray}
which explicitly takes into account the uncertainty over $f$. Hence, pure exploration of duels space can be carried out by maximizing
$$\alpha_{\pe}([\inputVector,\inputVector']| \dataSet_j) = \V[ \sigma(f_{\star}) | [\latentVector_{\star},\latentVector'_{\star}] | \dataSet_j ] .$$
Remark that in this case, duels that have been already visited will have a lower chance of being visited again even in cases in which the objective takes similar values in both players. See Figure~\ref{fig:exploration} for an illustration of this property.

In practice, this acquisition requires to compute and intractable integral, that we approximate in practice using Monte-Carlo.

\subsection{Copeland Expected Improvement}\label{subsec:CEI}
An alternative way to define an acquisition function is by generalizing the idea of the Expected Improvement \cite{Mockus77}. The idea of the \ei is to compute, in expectation, the marginal gain with respect to the current best observed output. In our context, as we do not have direct access to the objective, our only way of evaluating the quality of a single point is by computing its Copeland score. To generalize the idea to our context we need to find a couple of duels able to maximally improve the expected score of the Condorcer winner.

Denote by $c^{\star}_j$ the value of the Condorcet's winner when $j$ duels have been already run. For any new proposed duel $[\inputVector,\inputVector']$, two outcomes $\{0,1\}$ are possible that correspond to cases wether $\inputVector$ or $\inputVector'$ wins the duel. We denote by the $c^{\star}_{\inputVector,j}$ the value of the estimated  Condorcer winner resulting of augmenting $\dataSet$ with $\{[\inputVector,\inputVector'],1\}$ and by  $c^{\star}_{\inputVector',j}$ the equivalent value but augmenting the dataset with $\{[\inputVector,\inputVector'],0\}$. 
We define the one-lookahead Copeland Expected Improvement at iteration $j$ as:
\begin{equation}\label{eq:cei}
\alpha_{CEI}([\inputVector,\inputVector']| \dataSet_j) = \E \left[ (0, c - c_{j}^{\star})_{+}  | \dataSet_j\right]
\end{equation}
where  $(\cdot)_+ = \max(0,\cdot)$ and the expectation is take over $c$, the value at the Condorcet winner given the result of the duel. The next duel is selected as the pair that maximizes the \cei. Intuitively, the \cei evaluated at  $[\inputVector,\inputVector']$ is a weighted sum of the total increase of the best possible value of the Copeland score in the two possible outcomes of the duel. The weights are chosen to be the probability of the two outcomes, which are given by $\pi_{f}$. The \cei can be computed in closed form as 
\begin{eqnarray}\nonumber
\alpha_{CEI}([\inputVector,\inputVector']| \dataSet_j)  &= & \pi_{f}([\inputVector,\inputVector']|\dataSet_j)(c^{\star}_{\inputVector,j} -c_{j}^{\star})_{+}\\ \nonumber
&+ &\pi_{f}([\inputVector',\inputVector] |\dataSet_j)(c^{\star}_{\inputVector',j}-c_{j}^{\star})_{+}
\end{eqnarray}
The computation of this acquisition is computationally demanding as it requires updating of the \gp classification model for every fantasized output at any point in the domain.  As we show in the experimental section, and similarly with what is observed in the literature about the \ei, this acquisition tends to be greedy in certain scenarios leading to over exploitation \citep{Lobato2014,HennigSchuler2012}. Although non-myopic generalizations of this acquisition are possible to address this issue in the same fashion as in \citep{Gonzalez16glasses} these are be intractable. 

\subsection{Dueling-Thompson sampling}\label{subsec:duellingThomson}
As we have previously detailed, pure explorative approaches that do not exploit the available knowledge about the current best location and \cei is expensive to compute and tends to over-exploit. In this section we propose an alternative acquisition function that is fast to compute and explicitly balances \emph{exploration} and \emph{exploitation}. We call this acquisition dueling-Thompson sampling and it is inspired by Thompson sampling approaches.  It follows a two-step policy for duels:

\begin{figure*}[t!]
\centering
\includegraphics[width=0.335\textwidth]{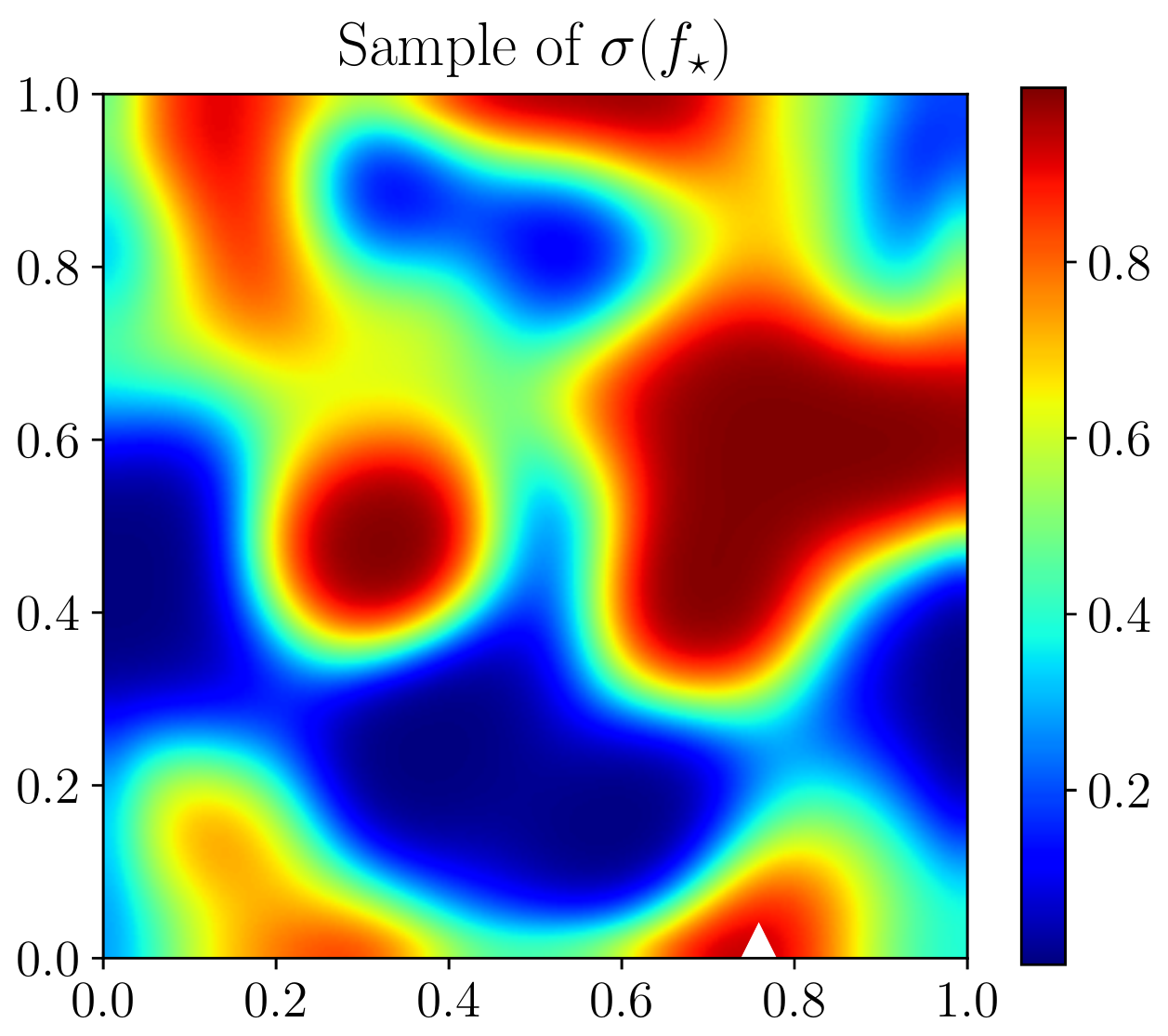} 
\includegraphics[width=0.295\textwidth]{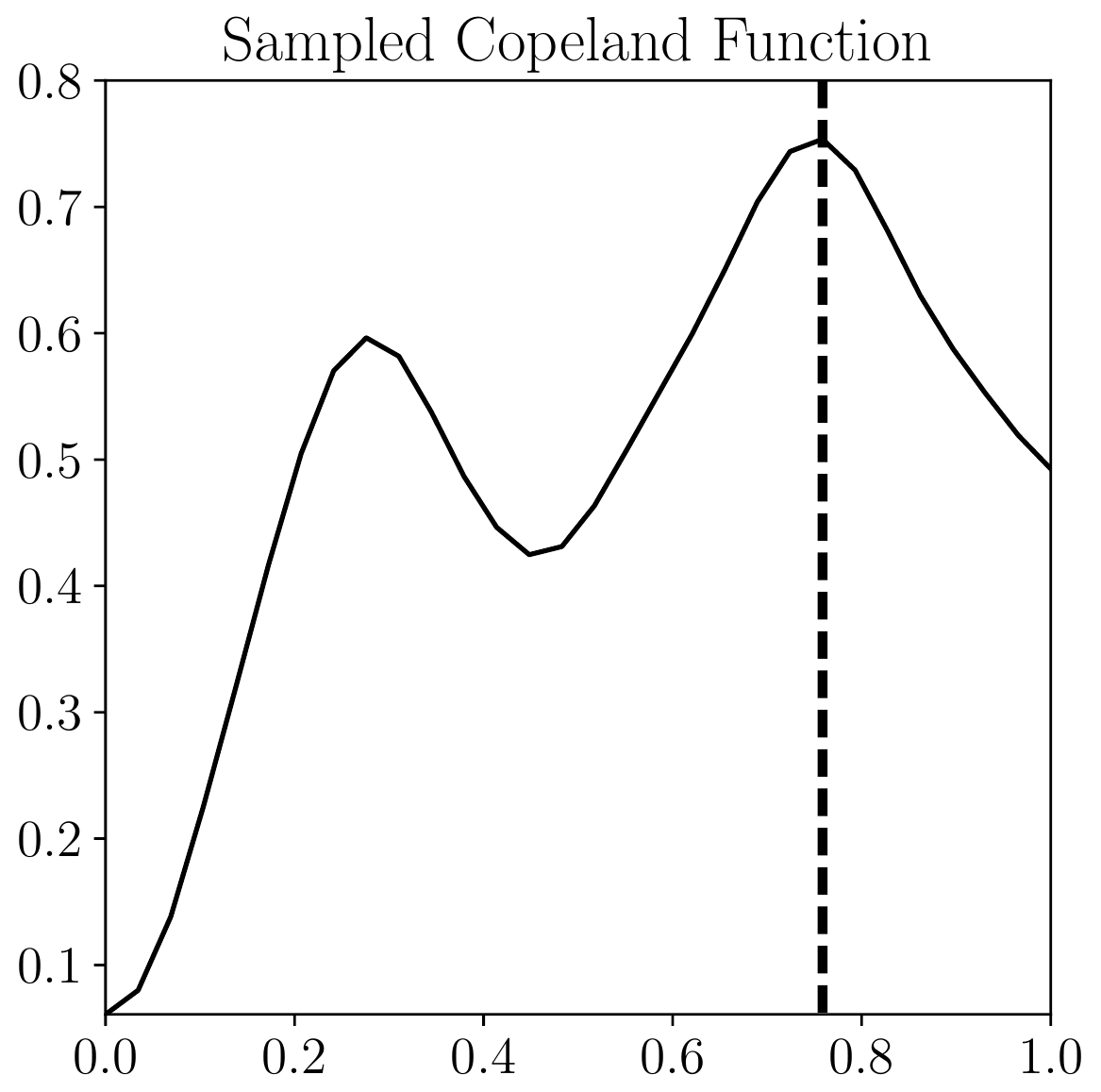} 
\includegraphics[width=0.34\textwidth]{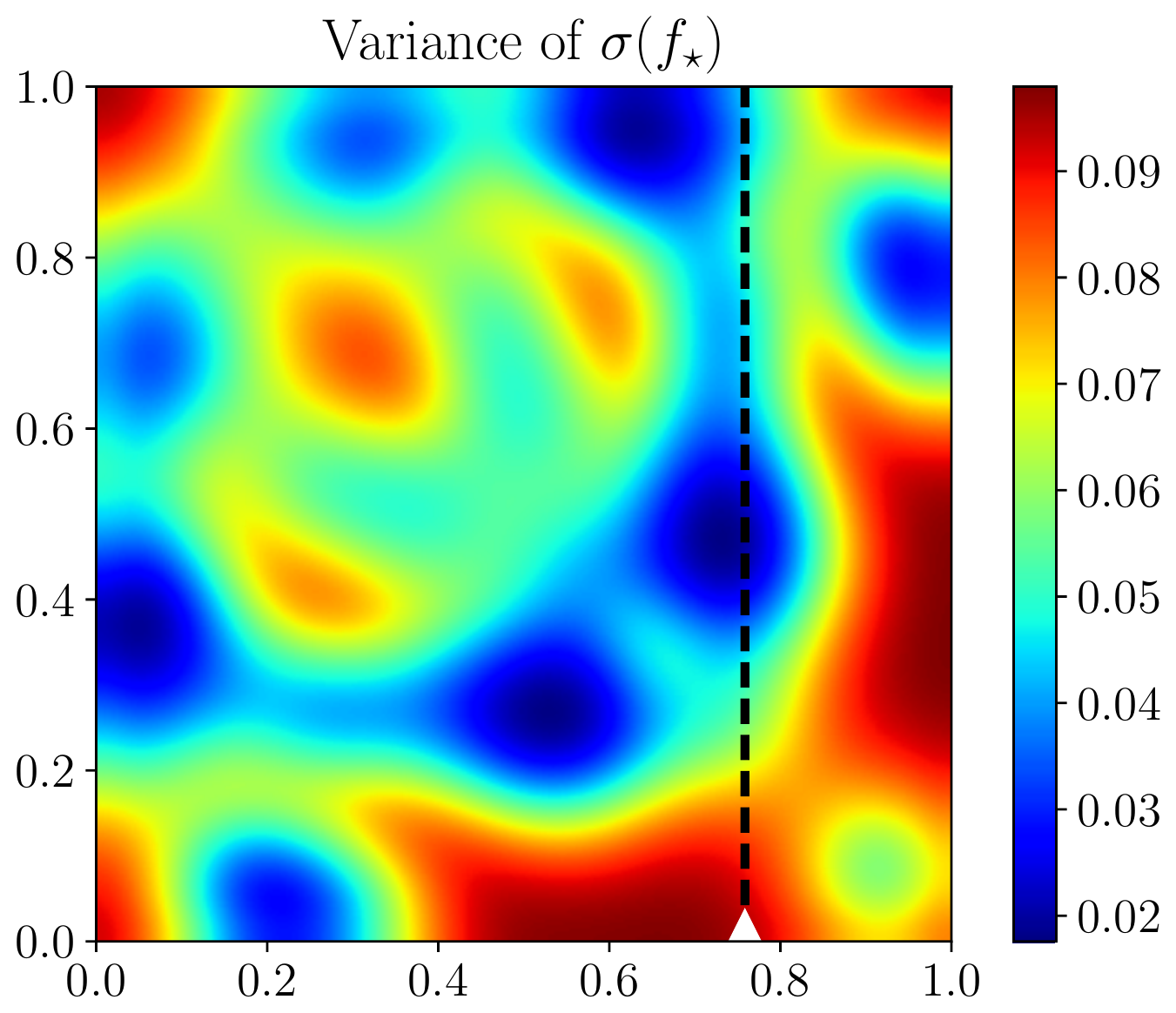} 
\vspace{-0.5cm}
\caption{Illustration of the steps to propose a new duel using the duelling-Thompson acquisition. The duel is computed using the same model as in Figure \ref{fig:exploration}. The white triangle represents the final selected duel. \emph{Left:} Sample from $f_{\star}$ squashed through the logistic function $\sigma$. \emph{Center:} Sampled soft-Copeland function, which results from integrating the the sample from $\sigma(f_{\star})$ on the left across the vertical axis. The first element of the duel $\inputVector$ is selected as the location of the maximum of the sampled soft-Copeland function (vertical dotted line). \emph{Right:} The second element of the duel, $\inputVector'$, is selected by maximizing the variance of $\sigma(f_{\star})$ marginally given $\inputVector$ (maximum across the vertical dotted line).}\label{fig:steps}
\end{figure*}

\begin{enumerate}
\item \emph{Step 1, selecting $\inputVector$:} First, generate a sample $\tilde{f}$ from the model using continuous Thompson sampling \footnote{Approximated continuous samples from a \gp with shift-invariant kernel can be obtained by using Bochner's theorem \cite{Bochner1959}. In a nutshell, the idea is to approximate the kernel by means of the inner product of a finite number Fourier features that are sampled according to the measure associated to the kernel \cite{Rahimi2008,Lobato2014}.} and compute the associated soft-Copland's score by integrating over $\mathcal{X}$. The first element of the new duel, $\inputVector_{next}$, is selected as:
$$\inputVector_{next} = \arg \max_{\inputVector \in \mathcal{X}} \int_{\mathcal X} \pi_{\tilde{f}}([\inputVector,\inputVector']; \dataSet_j) d\inputVector'.$$ 
The term $\text{Vol} ({\mathcal X})^{-1}$ in the Copeland score has been dropped here as it does not change the location of the optimum. The goal of this step is to balance exploration and exploitation in the selection of the Condorcet winner, it is the same fashion Thompson sampling does: it is likely to select a point close to the current Condorcet winner but the policy also allows exploration of other locations as we base our decision on a stochastic $\tilde{f}$. Also, the more evaluations are collected, the more greedy the selection will be towards the Condorcet winner. See Figure \ref{fig:samples} for an illustration of this effect.
\item \emph{Step 2, selecting $\inputVector'$:} Given $\inputVector_{next}$ the second element of the duel is selected as the location that maximizes the variance of $\sigma(f_{\star})$ in the direction of $\inputVector_{next}$. More formally, $\inputVector_{next}'$ is selected as
$$\inputVector_{next}' = \arg \max_{\inputVector'_{\star} \in \mathcal{X}} \V[ \sigma(f_{\star}) | [\latentVector_{\star},\latentVector'_{\star}] , \dataSet_j, \latentVector_{\star}=\inputVector_{next} ] $$
This second step is purely explorative in the direction of $x_{new}$ and its goal is to find informative comparisons to run with current good locations identified in the previous step. 
\end{enumerate}
In summary the dueling-Thompson sampling approach selects the next duel as:
$$\arg\max_{[\inputVector,\inputVector']} \alpha_{DTS}([\inputVector,\inputVector']| \dataSet_j) = [\inputVector_{next}, \inputVector_{next}'] $$
where $\inputVector_{next}$ and $\inputVector_{next}$ are defined above. This policy combines a selection of a point with high chances of being the optimum with a point whose result of the duel is uncertain with respect of the previous one. See Figure \ref{fig:steps} for a visual illustration of the two steps in toy example. See Algorithm \ref{alg:bopper} for a full description of the \us approach.

\begin{algorithm}[t!]
   \caption{The \us algorithm.}\label{alg:bopper}
   \label{alg:dbo}
\begin{algorithmic}
   \STATE {\bfseries Input:} Dataset $\dataSet_0 = \{[\latentVector_i,\latentVector'_i], \dataScalar_i\}_{i=1}^N$ and number of remaining evaluations $n$, acquisition for duels $\alpha([\inputVector,\inputVector'])$.
   \FOR{$j=0$ {\bfseries to} $n$ }
   \STATE 1. Fit a \gp with kernel $k$ to $\dataSet_{j}$ and learn $\pi_{f,j}(\latentVector)$.
   \STATE 2. Compute the acquisition for duels $\alpha$.
   \STATE 3. Next duel: $[\latentVector_{j+1},\latentVector'_{j+1}]= \arg \max\alpha([\inputVector,\inputVector'])$.
   \STATE 4. Run the duel $[\latentVector_{j+1},\latentVector'_{j+1}]$ and obtain $\dataScalar_{j+1}$.
   \STATE 5. Augment  $\dataSet_{j+1} = \{\dataSet_{j} \cup ([\latentVector_{j+1},\latentVector'_{j+1}], \dataScalar_{j+1})\}$.
   \ENDFOR
   \STATE Fit a \gp with kernel $\kernelScalar$ to $\dataSet_{n}$.
   \STATE \textbf{Returns}: Report the current Condorcet's winner $\latentVector^{\star}_n$.
\end{algorithmic}
\end{algorithm}

\begin{figure*}[t!]
\centering
\includegraphics[width=0.49\textwidth]{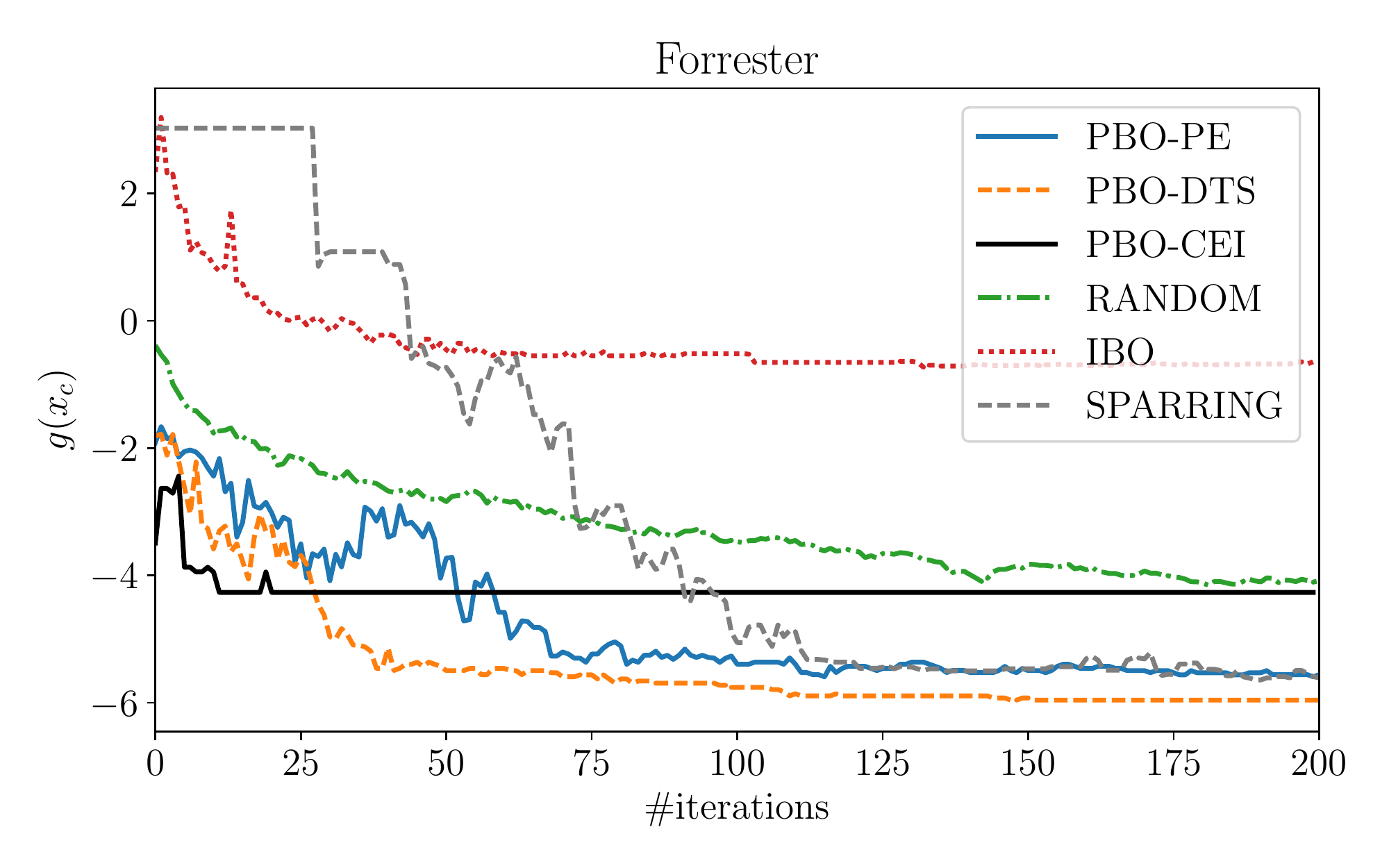} 
\includegraphics[width=0.49\textwidth]{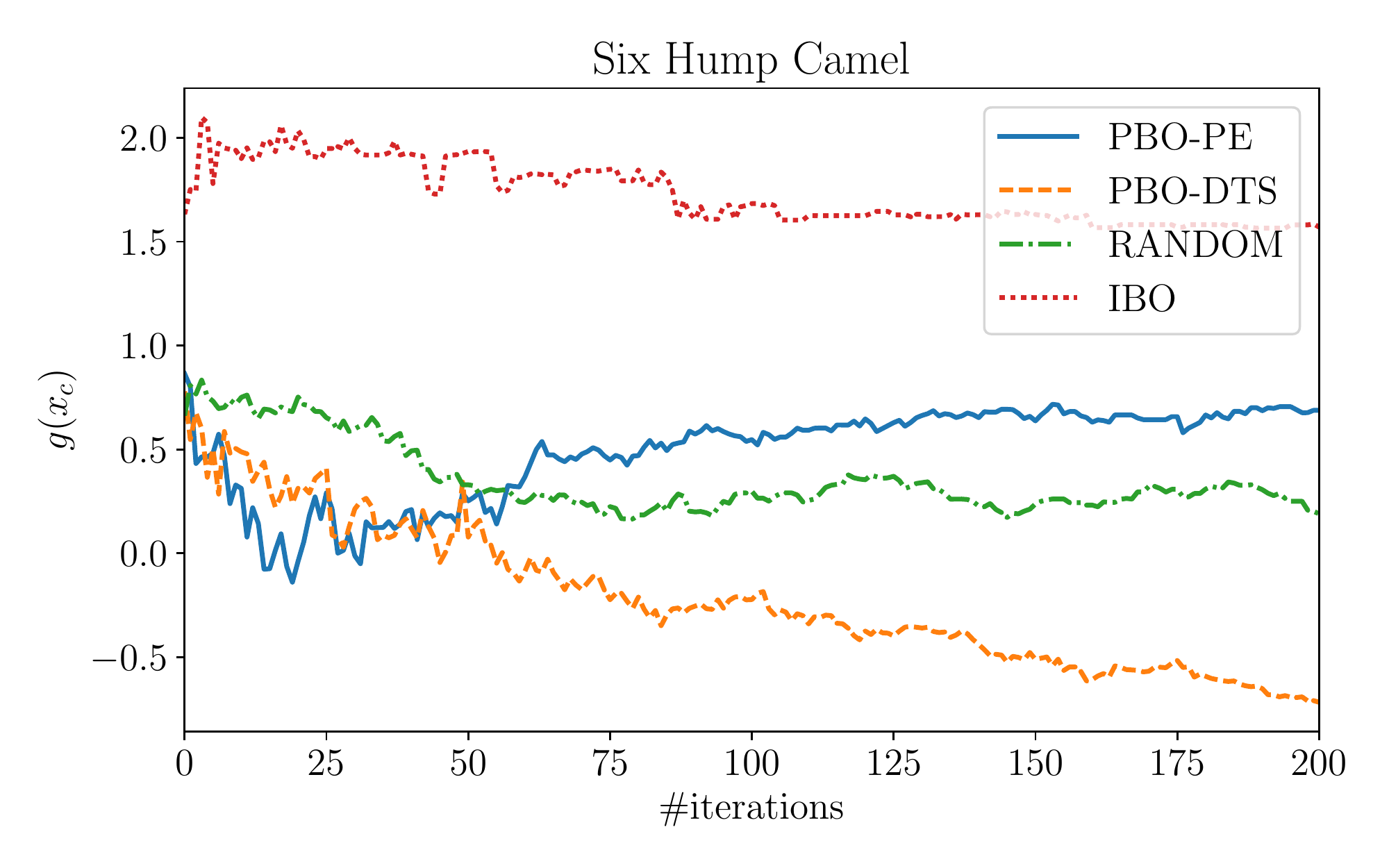} 
\includegraphics[width=0.49\textwidth]{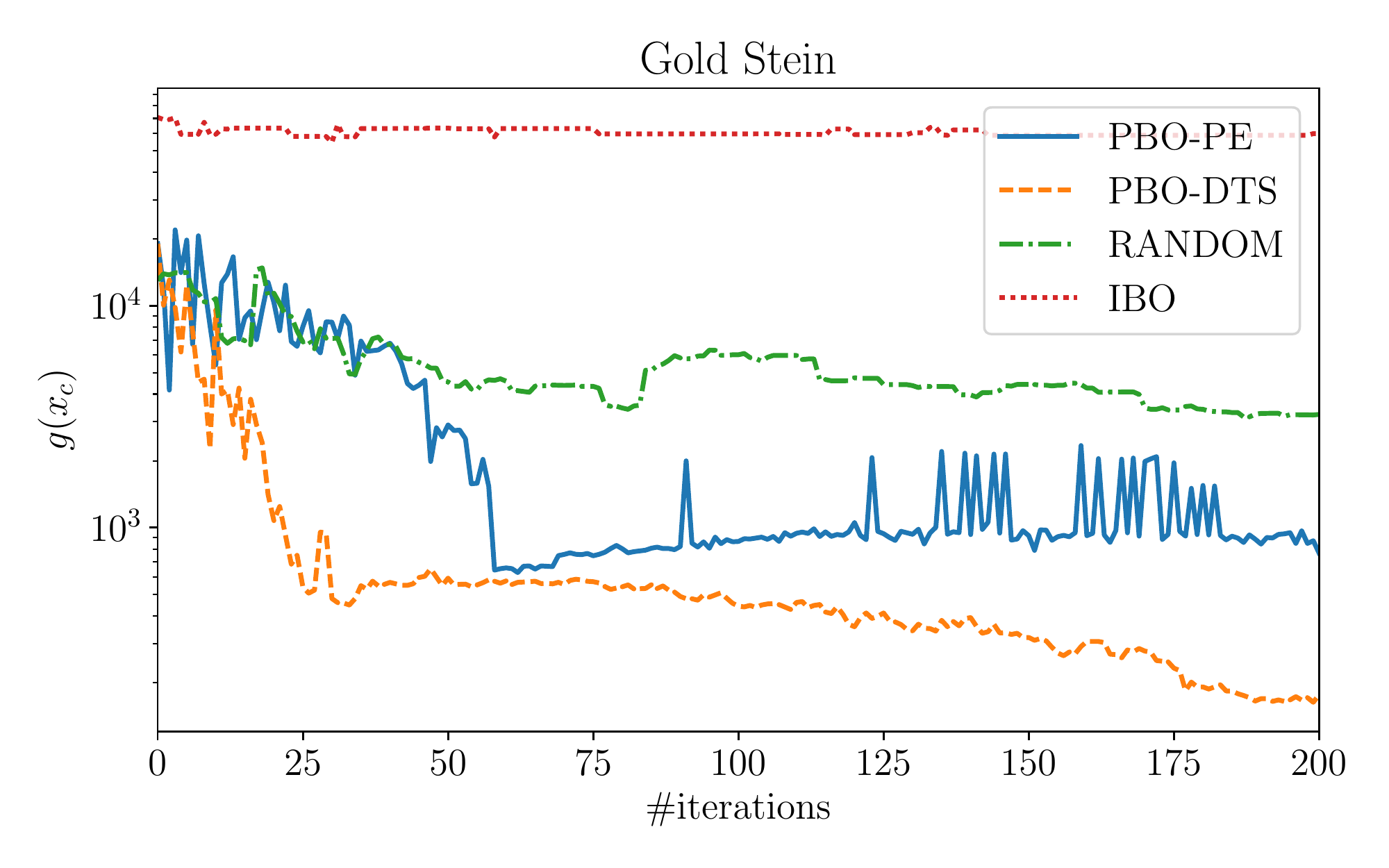} 
\includegraphics[width=0.49\textwidth]{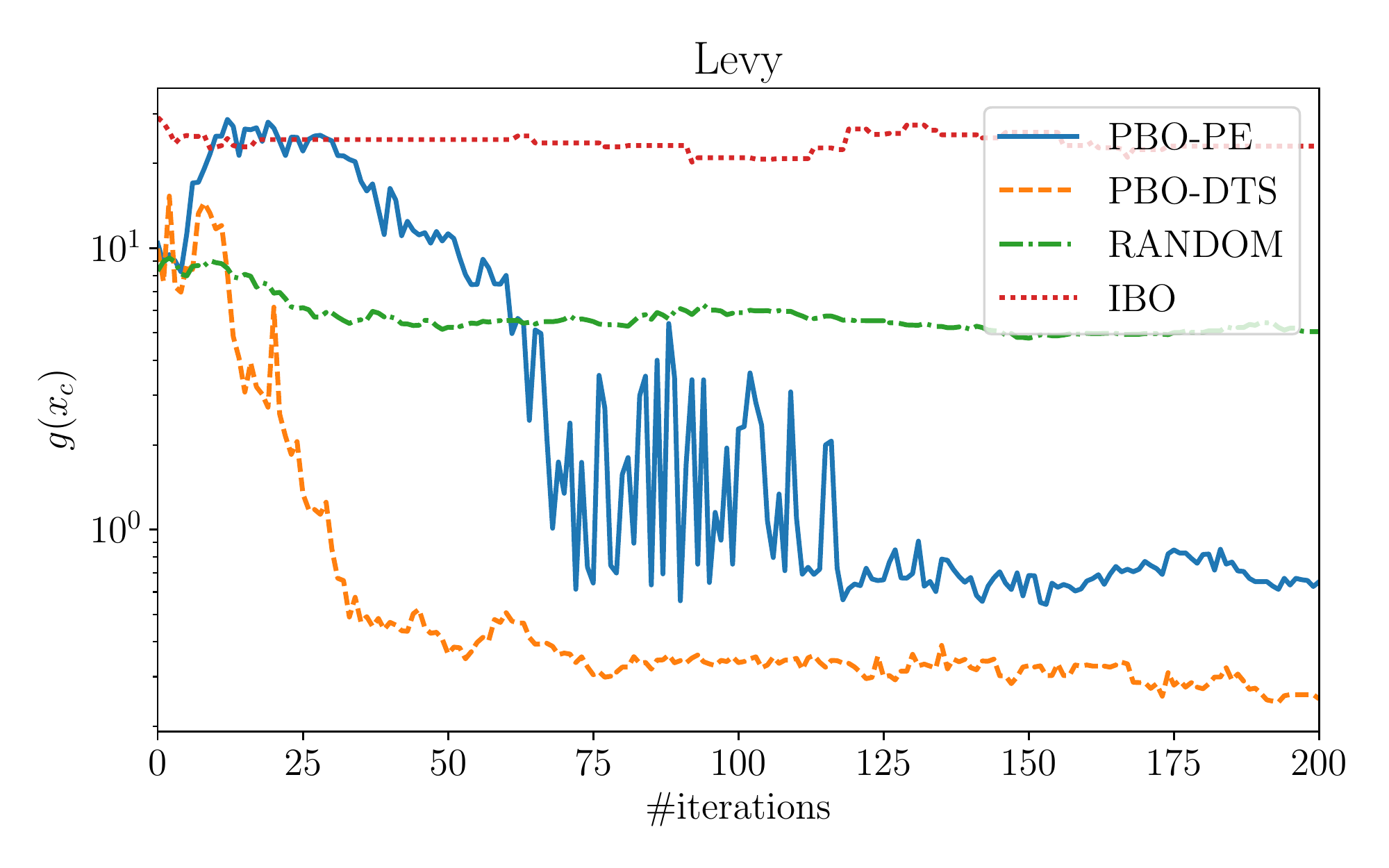} 
\vspace{-0.5cm}
\caption{Averaged results across 4 latent objective functions and 6 different methods. The \cei is only computed in the Forrester function as it is intractable when the dimension increases. Results are computed over 20 replicates in which 5 randomly selected initial points are used to initialize the models and that are kept the same across the six methods. The horizontal axis of the plots represents the number of evaluation and the vertical axis represent the value (log scale in the second row) of the true objective function at the best guess (Condorcet winner) at each evaluation. Note that this is only possible as we know the true objective function.The curves are not necessarily monotonically decreasing as we do not show the current best solution but the current solution at each iteration (proposed location by each method at each step in a real scenario). }\label{fig:experiments}
\end{figure*}

\subsection{Generalizations to multiple returns scenarios}

A natural extension of the \us set-up detailed above are cases in which multiple comparisons of inputs are simultaneously allowed. This is equivalent to providing a ranking over a set of points $\latentVector_1,\dots,\latentVector_k$. Rankings are trivial to map to pairwise preferences by using the pairwise ordering to obtain the result of the duels. The problem is, therefore, equivalent from a modeling perspective. However, from the point of view of selecting the $k$ locations to rank in each iteration, generalization of the above mentioned acquisitions are required. Although this is not the goal of this work, it is interesting to remark that this problem has strong connections with the one of computing batches in \bo \citep{GonDaiHenLaw16}.  

\section{Experiments}\label{sec:experiments}
\begin{figure}[t!]
\centering
\includegraphics[width=0.49\textwidth]{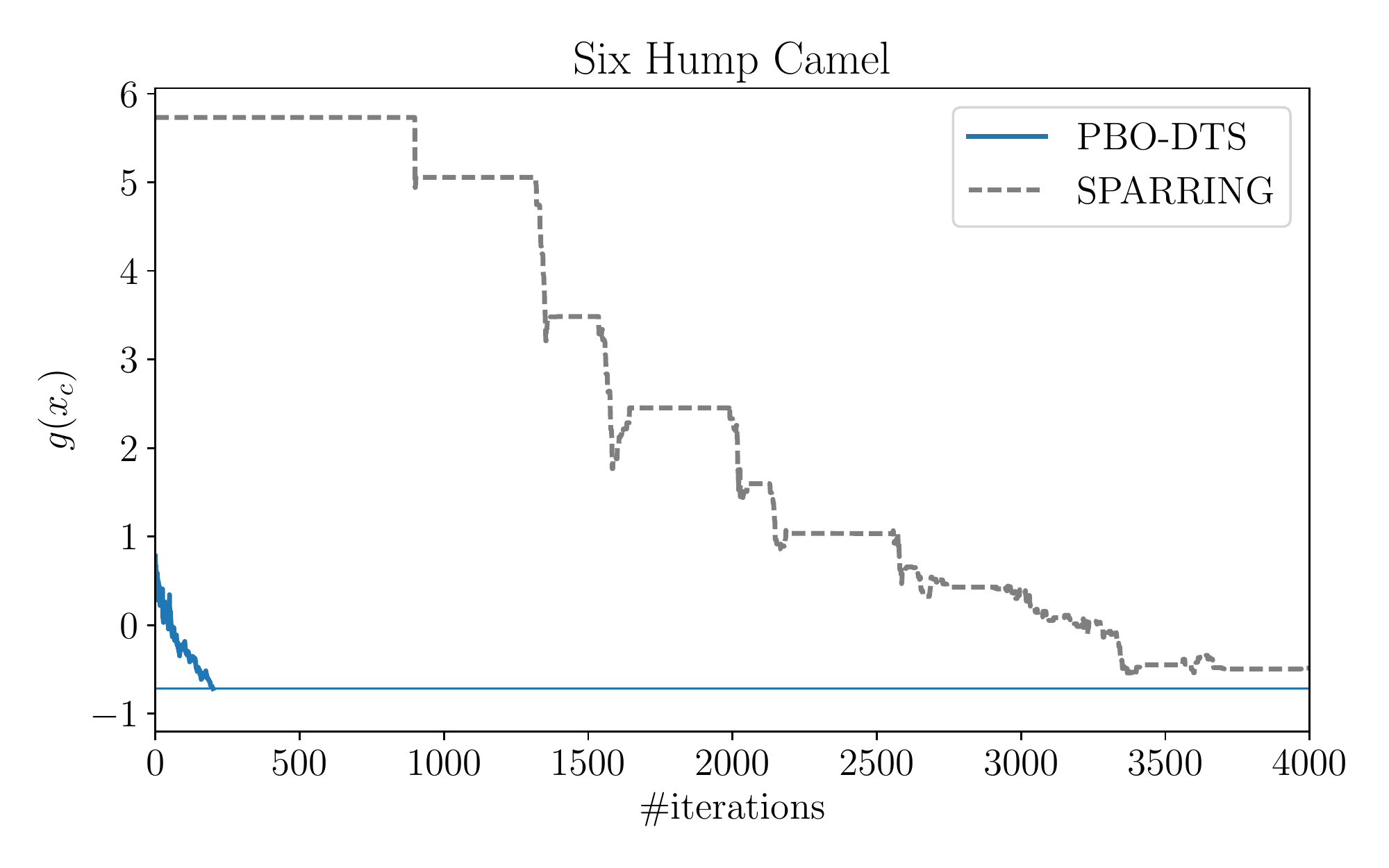} 
\vspace{-0.5cm}
\caption{Comparison of the bandits Sparring algorithm and \us-\dts on the Six-Hump Camel for an horizon in which the bandit method is run for 4000 iterations. The horizontal line represents the solution proposed by \us-\dts after 200 duels. As the plot illustrates, modeling correlations across the 900 arms (points in 2D) with a \gp drastically improves the speed at which the optimum of the function is found. Note that the bandit method needs to visit at least each arm before starting to get any improvement while \us-\dts is able to make a good (initial) guess with the first 5 random duels used in the experiment. Similar results are obtained for the rest of functions.}\label{fig:dts_sparring}
\end{figure}

In this section we present three experiments which validate our approach in terms of performance and illustrate its key properties. The set-up is as follows: we have a non-linear black-box function $g(\mathbf{x})$ of which we look for its minimum as described in equation \eqref{eq:min_problem}. However, we can only query this function through pairwise comparisons. The outcome of a pairwise comparison is generated as described in Section \ref{sec:approach}, i.e., the outcome is drawn from a Bernoulli distribution of which the sample probability is computed according to equation (\ref{eq:latent_probability}).

We have considered for $g(\mathbf{x})$ the Forrester, the `six-hump camel', `Gold-Stein' and `Levy' as latent objective functions to optimize. The Forrester function is 1-dimensional, whereas the rest are defined in domains of dimension 2. The explicit formulation of these objectives and the domains in which they are optimized are available as part of standard optimization benchmarks\footnote{https://www.sfu.ca/ssurjano/optimisation.html}. The \us framework is applicable in the continuous setting. In this section, however, the search of the optimum of the objectives is performed in a grid of size ($33$ per dimension for all cases), which has practical advantages: the integral in eq. (\ref{eq:integral}) can easily be treated as a sum and, more importantly, we can compare \us with bandit methods that are only defined in discrete domains. Each comparison starts with 5 initial (randomly selected) duels and a total budget of 200 duels are run, after which, the best location of the optimum should be reported. Further, each algorithm runs for 20 times (trials) with different initial duels (the same for all methods)\footnote{\random runs for 100 trials to give a reliable curve and \mbox{\us-\cei} only runs 5 trials for Forrester as it is very slow.}. We report the average performance across all trials, which is defined as the value of $g$ (known in all cases) evaluated at the current Condorcet winner, $\inputVector_c$ considered to be the best by each algorithm at each one of the 200 steps taken until the end of the budget. Note that each algorithm chooses $\inputVector_c$ differently, which leads to different performance at step 0. Therefore, we present plots of \#iterations versus $g(\inputVector_c)$.

We compare 6 methods. Firstly, the three variants within the \us framework: \us with pure exploration (\us-\pe, see section \ref{subsec:pureExplore}), \us with the Copeland Expected Improvement (\us-\cei, see section \ref{subsec:CEI}) and \us with dueling Thomson sampling (\us-\dts, see section \ref{subsec:duellingThomson}). We also compare against a random policy where duels are drawn from a uniform distribution (\random) \footnote{$\inputVector_c$ is chosen as the location that wins most frequently.} and with the interactive Bayesian optimization (\ibo) method of \cite{Brochu:2010}. \ibo selects duels by using an extension of Expected Improvement on a \gp model that encodes the information of the preferential returns in the likelihood. Finally, we compared against all three cardinal bandit algorithms proposed by \citet{AilonKJ14}, namely \textit{Doubler}, \textit{MultiSBM} and \textit{Sparring}. \citet{AilonKJ14} observes that \textit{Sparring} has the best performance, and it also outperforms the other two bandit-based algorithms in our experiments. Therefore, we only report the performance for \textit{Sparring} here, to keep the plots clean. In a nutshell, the \textit{Sparring} considers two bandit players (agents), one for each element of the duel, which use the Upper Confidence Bound criterion and where the input grid is acting as the set of arms. The decision for which pair of arms to query is according to the strategies and beliefs of each agent. Notice that, in this case, correlations in $f$ are not captured.

Figure \ref{fig:experiments} shows the performance of the compared methods, which is consistent across the four plots: \ibo shows a poor performance, due to the combination of the greedy nature of the acquisitions and the poor calibration of the model. The \random policy converges to a sub-optimal result and the \us approaches tend to be the superior ones. In particular, we observe that \us-\dts is consistently proven as the best policy, and it is able to balance correctly exploration and exploitation in the duels space. Contrarily, \us-\cei, which is only used in the first experiment due to the excessive computational overload, tends to over exploit. \us-\pe obtains reasonable results but tends to work worse in larger dimensions where is harder to cover the space.

Regarding the bandits methods, they need a much larger number of evaluations to converge compared to methods that model correlations across the arms. They are also heavily affected by an increase of the dimension (number of arms). The results of the \textit{Sparring} method are shown for the Forrester function but are omitted in the rest of the plots (the number of used evaluations used is smaller than the numbers of the arms and therefore no real learning can happen within the budget). However, in Figure \ref{fig:dts_sparring} we show the comparison between \textit{Sparring} and \us-\dts for an horizon in which the bandit method has almost converged. The gain obtained by modeling correlations among the duels is evident.

\section{Conclusions}\label{sec:conclusions}

In this work we have explored a new framework, \us, for optimizing black-box functions in which only preferential returns are available. The fundamental idea is to model comparisons of pairs of points with a Gaussian, which leads to the definition of new policies for augmenting the available dataset.
We have proposed three acquisitions for duels, \pe, \cei and \dts, and explored their connections with existing policies in standard \bo. Via simulation, we have demonstrated the superior performance of \dts, both because it finds a good balance between exploration and exploitation in the duels space and because it is computationally tractable. In comparison with other alternatives out of the \us framework, such as \ibo or other bandit methods, \dts shows the state-of-the-art performance.

We envisage some interesting future extensions of our current approach. The first one is to tackle the existing limitation on the dimension of the input space, which is doubled with respect to the original dimensionality of the problem to be able to capture correlations among duels. Also further theoretical analysis will be carried out on the proposed acquisitions.

\bibliographystyle{plainnat}
\bibliography{bib_bopper}

\end{document}